\documentclass[runningheads]{llncs}

\usepackage{eccv}

\usepackage{eccvabbrv}

\usepackage{graphicx}
\usepackage{booktabs}
\usepackage{colortbl}
\usepackage{subcaption}
\usepackage[accsupp]{axessibility}  

\usepackage{hyperref}
\usepackage{wrapfig}
\usepackage{orcidlink}

\begin{document}

\title{Fractal Autoregressive Depth Estimation with Continuous Token Diffusion} 

\titlerunning{Abbreviated paper title}

\author{Jinchang Zhang\inst{1} \and
Xinrou Kang\inst{1} \and
Guoyu Lu\inst{1}}

\authorrunning{F.~Author et al.}

\institute{Intelligent Vision and Sensing (IVS) Lab at SUNY Binghamton University
\email{guoyulu62@gmail.com}}

\maketitle

\begin{abstract}
  Monocular depth estimation can benefit from autoregressive (AR) generation, but direct AR modeling is hindered by RGB–depth modality mismatch, inefficient pixel-wise generation, and instability in continuous depth prediction. We propose a Fractal Visual Autoregressive Diffusion framework that reformulates depth estimation as a coarse-to-fine next-scale autoregressive generation process. A VCFR module fuses multi-scale image features with current depth predictions to improve cross-modal conditioning, while a conditional denoising diffusion loss models depth distributions directly in continuous space and mitigates discrete quantization error. To improve computational efficiency, we organize the scale-wise generators into a fractal recursive architecture, reusing a base visual AR unit in a self-similar hierarchy. We further introduce an uncertainty-aware robust consensus aggregation scheme for multi-sample inference to improve fusion stability and provide a practical pixel-wise reliability proxy. Experiments on standard benchmarks show strong performance and validate the proposed design.
\end{abstract}

\section{Introduction}
Monocular Depth Estimation is a fundamental task in robotic perception and computer vision, and plays an important role in applications such as scene understanding, 3D reconstruction, and autonomous navigation~\cite{izadi2011kinectfusion,chen2019towards,wang2019pseudo}. Its goal is to predict pixel-wise depth from a single RGB image.
Autoregressive (AR) generative models have demonstrated strong sequential modeling capability in image generation, and they also provide a potential path for decomposing complex depth prediction into structured subproblems. However, directly applying conventional AR models to monocular depth estimation still faces three major challenges: (1) cross-modal mapping makes it difficult to model structural correspondences between RGB appearance and depth geometry; (2) pixel-wise generation leads to high serial inference cost; and (3) error propagation during generation undermines prediction stability and reliability.

To address these issues, we propose an autoregressive depth generation framework based on next-scale prediction, which reformulates monocular depth estimation as a progressive generation process from low to high resolution and uses scale, rather than pixel, as the generation unit. This design improves inference efficiency while preserving hierarchical modeling capacity.
To alleviate the modality mismatch between RGB and depth, we design a Visual-Conditioned Feature Refinement (VCFR) module. This module constructs a Visual-Depth Joint token as the conditional input for next-scale depth generation, thereby enhancing cross-modal structural alignment and detail recovery.
In addition, another major challenge of traditional AR models lies in the quantization process. Discretizing continuous depth values into tokens often introduces quantization error, causing fine-grained geometric information loss and unstable training. However, the core principle of autoregressive modeling—predicting the next state conditioned on the existing context—does not inherently require a discrete state space~\cite{li2024autoregressive}. Based on this observation, while retaining the scale-wise AR generation framework, we incorporate a conditional denoising diffusion loss to model the conditional distribution at each scale in continuous depth space. This design avoids quantization errors caused by discrete tokenization and provides a more flexible conditional distribution representation than point-wise regression in locally ambiguous regions. Specifically, the autoregressive process models cross-scale hierarchical dependencies, while the diffusion process models the continuous depth distribution within each scale; the two are complementary.
However, merely reformulating depth estimation as scale-wise progressive generation is still insufficient to  resolve the computational burden introduced by multi-scale iteration. If different scales use independent generators, both model complexity and inference path length increase. To this end, we further propose a fractal architecture, which abstracts the scale-wise autoregressive unit as a reusable base generator and organizes the overall model in a self-similar recursive manner. This design structurally unifies the generation process across scales, aligning the modeling principle of next-scale prediction with the network implementation, and mitigates the growth of parameters and inference cost through module reuse. As a result, it achieves a better efficiency-performance trade-off while maintaining modeling capacity.
Building on this, to address the instability and limited reliability of AR-generated depth during inference, we further propose an Uncertainty-Aware Robust Consensus Aggregation module for stable fusion of multiple sampled predictions and reliability estimation. This design improves the stability of multi-sample aggregation without additional training and provides practical region-level risk indicators for downstream tasks.

\begin{figure*}[t]
\begin{center}
\includegraphics[width=12cm, height=5cm]{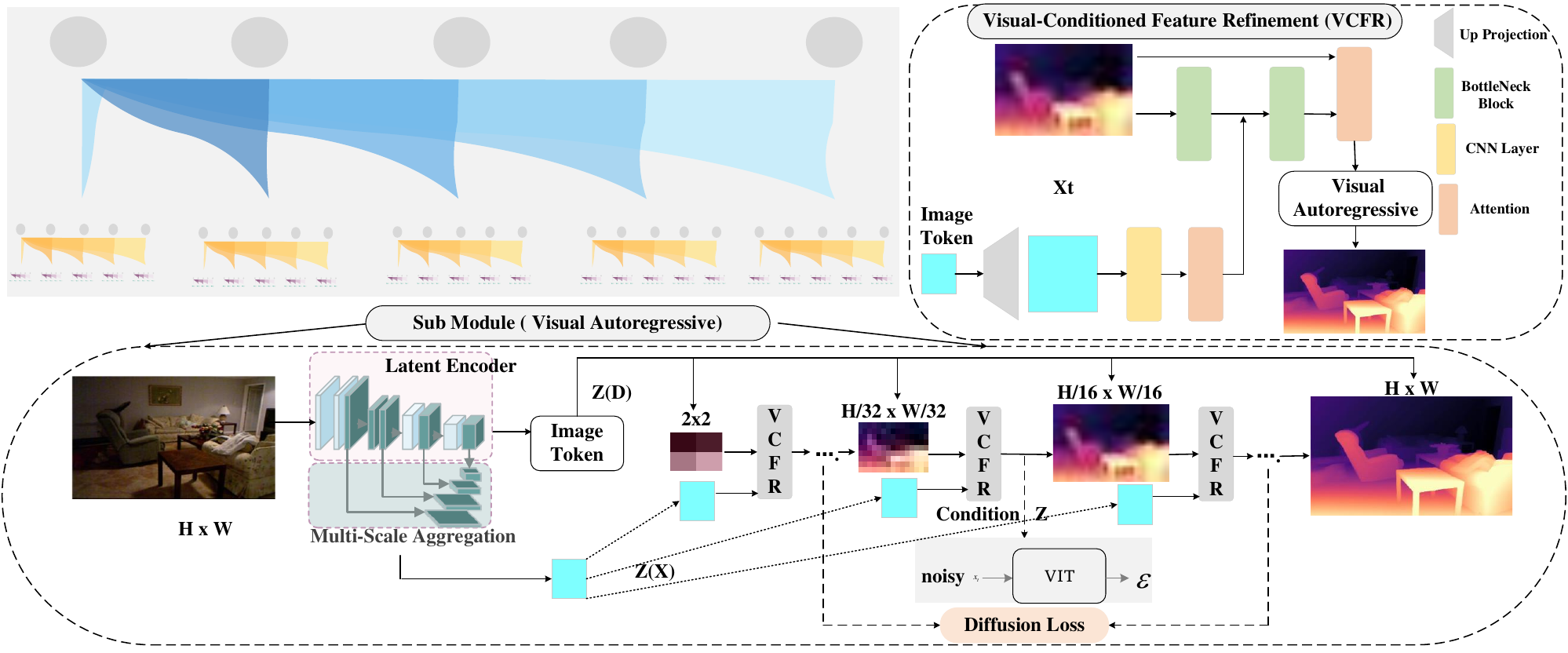}
\end{center}
\vspace{-4mm}
\caption{
Given an input RGB image $X \in \mathbb{R}^{H \times W \times 3}$, we first extract multi-scale visual features $Z(X)$ using an image encoder and a multi-scale aggregation module, which serve as global conditioning throughout depth generation. The framework is organized as a fractal recursive hierarchy of four nested scale-wise generators $\{g_4, g_3, g_2, g_1\}$, which progressively recover depth representations from coarse to fine and output the final depth map $\hat{D} \in \mathbb{R}^{H \times W}$. Specifically, $g_4$ starts from a coarse latent token and predicts an initial depth latent, which is then passed to $g_3$, followed by $g_2$, and finally $g_1$. The top-left panel illustrates the fractal recursion pattern, while the bottom panel shows the concrete generation pipeline. At each scale, a VCFR block fuses image features with the current-scale depth latent/token to form a Visual-Depth Joint token for next-scale prediction. Within each module, depth is predicted via a visual autoregressive diffusion process, where a conditional denoising diffusion objective models the continuous depth distribution at the current scale; the noise predictor $\epsilon_{\theta}(x_t \mid t, z)$ takes diffusion timestep $t$ and condition vector $z$ and is optimized with diffusion loss.
}
\vspace{-4mm}
\label{architecture}
\end{figure*}

We summarize our contributions as follows:
1.
We propose a fractal recursive autoregressive framework for monocular depth estimation, which reformulates depth prediction as a next-scale generation process and recursively composes scale-wise generators using reusable autoregressive modules.
2.
We introduce a VCFR module that constructs a Visual-Depth Joint token by fusing multi-scale image features with current depth predictions, thereby improving cross-modal conditioning during autoregressive generation.
3. 
We incorporate a conditional denoising diffusion loss into the scale-wise autoregressive process to directly model depth distributions in continuous space, reducing quantization errors caused by discrete tokenization.
4. We propose an uncertainty-aware robust consensus aggregation scheme for multi-sample inference, which improves fusion stability and provides a practical pixel-wise reliability proxy without additional training. Our framework is illustrated in Fig.~\ref{architecture}.

\section{Related Work}

\subsection{Monocular Depth Estimation}

\textbf{Supervised Learning: }
Supervised learning methods typically formulate monocular depth estimation as a per-pixel regression problem and train on datasets with ground-truth depth annotations. Early approaches were primarily CNN-based, such as RAP \cite{zhang2019pattern} and DAV \cite{huynh2020guiding}, which improved depth prediction through enhanced feature extraction and refinement modules. More recently, Transformer-based architectures have also been introduced, e.g., PixelFormer \cite{agarwal2023attention}. In addition, \cite{zhang2025language} proposes an image-fusion method better tailored to depth estimation, while \cite{zhang2025depth} explores depth estimation from defocus cues.
Another line of work formulates depth estimation as a hybrid regression--classification problem. Methods such as AdaBins \cite{bhat2021adabins} adopt adaptive binning strategies that combine classification and regression for depth prediction. WordDepth \cite{zeng2024wordepth} is built on a VAE framework and uses textual descriptions to model a depth distribution; image-conditioned samples are then combined with this distribution to infer depth. Building on this idea, \cite{zhang2024embodiment,zhang2025vision} further incorporates a camera model to compute depth priors from environmental information.

\subsection{Generative Models in Depth Estimation}

\textbf{Diffusion Models:}
Marigold \cite{ke2023repurposing} shows that pre-trained text-to-image diffusion generators can be effectively repurposed for monocular depth estimation. ECoDepth \cite{patni2024ecodepth} further highlights the importance of conditioning design by constructing more effective visual conditioning signals to improve the geometric prediction quality of diffusion-based depth estimation. DiffusionDepth \cite{duan2024diffusiondepth} progressively recovers depth maps from noisy depth states through iterative denoising. Lotus \cite{helotus} revisits diffusion design choices from a more general dense prediction perspective and proposes a simplified diffusion formulation that is better suited for dense prediction.
\textbf{Autoregressive Models:}
Another line of generative modeling for depth estimation is based on autoregressive ideas. DORN \cite{fu2018deep} estimates depth by discretizing the depth space into multiple bins and predicting the corresponding ordinal labels. Subsequent work, such as Ord2Seq \cite{wang2023ord2seq}, further formulates ordinal regression as a sequence generation problem and uses autoregressive networks to progressively refine predictions. \cite{wang2025scalable} formulates monocular depth estimation under an autoregressive prediction paradigm.
However, these methods still rely on discretization-based modeling pipelines, which may introduce information loss before depth decoding and limit the representation of fine-grained continuous geometry.
In contrast, our framework combines scale-wise visual autoregression with conditional diffusion modeling in a continuous latent space, avoiding explicit discretization while enabling coarse-to-fine recursive depth generation with improved computational efficiency.
\vspace{-1mm}
\section{Method}
\vspace{-1mm}
Given a monocular RGB image, the model first extracts multi-scale visual features through an image encoder. Subsequently, depth prediction is no longer performed via pixel-wise autoregressive generation; instead, it is decomposed into a coarse-to-fine cross-scale recursive generation process, which is progressively realized by a fractal hierarchical generator $(g_4 \rightarrow g_1)$. Specifically, $g_4$ first predicts the depth latent variable at the coarsest scale. Then, each subsequent generator takes the depth state from the previous level as a condition and generates a higher-precision continuous depth latent variable at the corresponding resolution, until $g_1$ outputs the highest-resolution depth representation. 
To alleviate the modality mismatch between RGB features and depth representations, we introduce a Visual-Condition Feature Refinement (VCFR) module at each level, which fuses the current depth latent variable with visual features at the same scale to construct a joint conditional representation for the next prediction step. The depth latent variables at all scales are modeled in a continuous space using a conditional diffusion objective. Finally, the highest-resolution depth latent variable is decoded into a dense depth map.
\vspace{-1mm}
\subsection{Preliminaries: Visual Autoregressive Modeling}
\label{sec:var_prelim}

We briefly review the visual autoregressive (VAR) formulation that serves as the foundation of our method, and adapt its notation to the depth prediction setting for completeness. 
Traditional autoregressive modeling assumes a 1D causal dependency, where each token depends only on its prefix. However, visual representations are naturally organized as 2D feature maps with strong bidirectional spatial correlations. Directly flattening a feature map into a 1D sequence not only breaks local spatial structure, but also introduces a mismatch between the spatially correlated visual tokens and the unidirectional dependency assumption of standard token-wise autoregressive models~\cite{tian2024visual}.
To address this issue, we follow the visual autoregressive modeling strategy in~\cite{tian2024visual} and shift the prediction unit from the next token to the next scale. Instead of autoregressively generating individual tokens in raster order, the model predicts a textbf{token map} at each step. Concretely, let $\mathbf{F} \in \mathbb{R}^{h \times w \times C}$ denote a visual token map, and let $(r_1, r_2, \dots, r_K)$ denote its multi-scale representations with progressively increasing spatial resolutions $(h_k \times w_k)$, where the final scale $r_K$ matches the original resolution $(h \times w)$. The scale-wise autoregressive factorization is written as
$
P(r_1, r_2, \dots, r_K) = \prod_{k=1}^{K} P(r_k \mid r_{<k}),
$
\noindent where $r_{<k} := (r_1, \dots, r_{k-1})$ denotes the prefix over scales. In this formulation, each autoregressive step predicts all tokens in $r_k$ in parallel conditioned on the previously generated coarser-scale token maps.
This scale-wise formulation defines an autoregressive dependency over textbf{scales} rather than over individual tokens, thereby preserving within-scale spatial structure and avoiding raster-order flattening during generation. In our method, we build on this factorization principle and further reformulate it into a recursive fractal framework for coarse-to-fine depth generation.
\vspace{-1mm}
\subsection{Fractal Autoregressive Framework}
\label{sec:fractal_ar}

While the scale-wise VAR formulation in Sec.~\ref{sec:var_prelim} preserves spatial structure and enables parallel generation within each scale, directly applying a monolithic visual autoregressive module to dense depth prediction remains computationally demanding. In particular, as the target resolution increases, the model must jointly capture increasingly complex geometric dependencies, which leads to high modeling and inference cost. To address this issue, we reformulate scale-wise autoregressive depth generation as a recursive fractal process.
Our key idea is to construct a hierarchy of self-similar generators that progressively refine depth from coarse to fine scales. Let $\mathbf{z}_i$ denote the depth latent (or depth state) at recursion level $i$, where lower levels correspond to coarser spatial resolutions and higher levels correspond to finer resolutions. Instead of using a single generator to predict the full-resolution depth representation in one step, we define a sequence of level-wise generators $\{g_i\}$, each responsible for refining the depth representation at the next scale:
$
\label{eq:fractal_recursion}
\mathbf{z}_{i+1} = g_i(\mathbf{z}_i, \mathbf{c}_i),
$
\noindent where $\mathbf{c}_i$ denotes a scale-specific conditioning signal derived from the input image and the current depth state (detailed in Sec.~\ref{sec:vcfr}). This recursive formulation decomposes dense depth generation into a sequence of manageable subproblems, each operating at a predefined spatial scale.
From a probabilistic perspective, the recursive process induces a hierarchical autoregressive factorization over depth representations across scales. Let $\mathbf{z}_{1:L} := (\mathbf{z}_1, \mathbf{z}_2, \dots, \mathbf{z}_L)$ denote the depth representations from the coarsest to the finest level. Conditioned on the input image $I$, we write
\vspace{-1mm}
\begin{equation}
\label{eq:fractal_factorization}
p(\mathbf{z}_{1:L} \mid I)
=
p(\mathbf{z}_1 \mid I)\prod_{i=1}^{L-1} p(\mathbf{z}_{i+1} \mid \mathbf{z}_i, \mathbf{c}_i),
\vspace{-2mm}
\end{equation}
\vspace{-1mm}
\noindent where each factor models the transition from one scale to the next finer scale. Compared with monolithic pixel-wise autoregressive modeling, this factorization provides a more favorable computational structure by distributing the generation burden across multiple levels, while explicitly aligning with the coarse-to-fine hierarchical nature of scene geometry.
We instantiate this recursive hierarchy as a fractal stack $g_4$ to $g_1$, where each module predicts a depth latent at a predefined resolution and passes it to the next finer level for further refinement. The visual-depth conditioning mechanism used to construct $\mathbf{c}_i$ is introduced next.

\vspace{-1mm}
\subsection{Visual-Conditioned Feature Refinement (VCRF)}
\label{sec:vcfr}

As introduced in Sec.~\ref{sec:fractal_ar}, each recursive generator $g_i$ requires a scale-specific conditioning signal $\mathbf{c}_i$ to inject image evidence during coarse-to-fine depth refinement. Relying only on the previous depth state $\mathbf{z}_i$ may be insufficient to recover fine-grained geometric structures, especially around object boundaries and texture-rich regions. To bridge this modality gap between RGB appearance and depth representation, we introduce a VCFR module that constructs a visual-depth joint condition at each scale.
\textbf{Multi-scale visual feature extraction.}
Given an input image $I$, we first extract multi-scale visual features using a Swin Transformer backbone~\cite{liu2021swin}, which captures both coarse scene layout and fine local details. We further enhance cross-scale feature interaction via the HAHI module~\cite{li2023depthformer}. The resulting multi-scale features are then aggregated with a  FPN~\cite{lin2017feature} to produce scale-aligned visual features $\{\mathbf{f}_i\}_{i=1}^{L}$ for recursive depth refinement.
\textbf{Scale-wise visual-depth refinement.}
At recursion level $i$, VCFR takes the current depth state $\mathbf{z}_i$ and the corresponding visual feature $\mathbf{f}_i$ as input, and outputs a refined conditioning feature $\mathbf{c}_i$. To preserve local structural consistency while matching spatial resolutions, we first apply a local projection layer to align $\mathbf{f}_i$ to the spatial size of $\mathbf{z}_i$. The projected visual feature is then processed by a lightweight refinement head consisting of a convolutional block and a self-attention layer, which jointly model local patterns and long-range dependencies. We then fuse the refined visual feature with the current depth state through element-wise interaction, followed by a bottleneck block~\cite{he2016deep} and a channel-attention module with residual connections for feature recalibration and stabilization. The final output is the scale-specific visual-depth joint condition $\mathbf{c}_i$, which is used by $g_i$ to predict the next-scale depth representation in Eq.~\eqref{eq:fractal_recursion}.
VCFR injects image structure into each recursive refinement stage, rather than relying on depth-only propagation across scales. This design improves cross-modal alignment between RGB features and depth latents while maintaining a lightweight conditioning path for efficient multi-level inference.
\subsection{Conditional Diffusion Modeling in Continuous  Latent Space}
\label{sec:continuous_token}

The scale-wise autoregressive factorization in Secs.~\ref{sec:var_prelim} and~\ref{sec:fractal_ar} specifies {how} depth representations are generated across scales, but does not by itself determine \emph{how} the per-scale conditional distribution should be modeled. Prior autoregressive formulations often rely on discrete tokenization, which can be effective for image generation but is less natural for depth estimation, where the target encodes continuous scene geometry. In particular, discretization may introduce quantization artifacts and can limit the fidelity of fine-grained depth variations. Following the view that autoregressive modeling fundamentally concerns distribution factorization rather than mandatory vector quantization~\cite{li2024autoregressive}, we model depth latents directly in continuous space.
Given the recursive depth state and the scale-specific visual-depth condition from Secs.~\ref{sec:fractal_ar} and~\ref{sec:vcfr}, we model the conditional distribution at each level using a conditional diffusion objective. Concretely, at scale $i$, let $\mathbf{z}_i^\ast$ denote the target depth latent (derived from ground-truth depth) and let $\mathbf{c}_i$ denote the VCFR conditioning feature. Our goal is to model the conditional distribution
$
p(\mathbf{z}_{i+1}^{*} \mid \mathbf{z}_i, \mathbf{c}_i),
$
\noindent where $\mathbf{c}_i$  encodes the autoregressive context through the current recursive state and image evidence.

\textbf{Conditional diffusion objective.}
We adopt a denoising diffusion objective~\cite{ho2020denoising} to supervise continuous depth latent modeling at each scale. Let $\epsilon \sim \mathcal{N}(0,\mathbf{I})$ and let $t$ be a diffusion timestep sampled from a predefined noise schedule. The noised latent is constructed as
$
\mathbf{z}_{i,t}
=
\sqrt{\bar{\alpha}_t}\,\mathbf{z}_i^\ast
+
\sqrt{1-\bar{\alpha}_t}\,\epsilon,
$
\noindent where $\bar{\alpha}_t$ is the cumulative product of the noise schedule coefficients. A conditional noise predictor $\epsilon_\theta$ takes $(\mathbf{z}_{i,t}, t, \mathbf{c}_i)$ as input and predicts the injected noise. The training loss is
$
\mathcal{L}_{\mathrm{diff}}^{(i)}
=
\mathbb{E}_{t,\epsilon}
\left[
\left\|
\epsilon - \epsilon_\theta(\mathbf{z}_{i,t}, t, \mathbf{c}_i)
\right\|_2^2
\right].
$
Compared with direct discrete token matching, this objective enables continuous distribution modeling of depth latents and better preserves smooth geometric transitions across scales.

\textbf{Relation to direct regression.}
A direct regression loss (e.g., $\ell_1/\ell_2$ or Gaussian likelihood regression) typically optimizes a point estimate under a relatively restrictive output assumption. In contrast, conditional diffusion models the full conditional distribution in continuous latent space, which provides a more expressive objective for handling local geometric ambiguity (e.g., occlusion boundaries and texture-poor regions) while maintaining compatibility with our scale-wise autoregressive recursion.

\textbf{Training and sampling.}
During training, diffusion timesteps are sampled randomly for each scale, and the same scale-specific condition $\mathbf{c}_i$ can be reused across sampled timesteps for efficient supervision. At inference, we sample the depth latent from the learned conditional distribution via the reverse diffusion process. Starting from Gaussian noise $\mathbf{z}_{i,T} \sim \mathcal{N}(0,\mathbf{I})$, we iteratively denoise using
\vspace{-1mm}
\begin{equation}
\label{eq:reverse_sampling}
\mathbf{z}_{i,t-1}
=
\frac{1}{\sqrt{\alpha_t}}
\left(
\mathbf{z}_{i,t}
-
\frac{1-\alpha_t}{\sqrt{1-\bar{\alpha}_t}}
\epsilon_\theta(\mathbf{z}_{i,t}, t, \mathbf{c}_i)
\right)
+
\sigma_t \delta,
\end{equation}
\vspace{-1mm}
\noindent where $\delta \sim \mathcal{N}(0,\mathbf{I})$ and $\sigma_t$ is the noise scale at timestep $t$. The final sample $\mathbf{z}_{i,0}$ is used as the refined depth latent at scale $i$. Optionally, we use a temperature parameter $\tau$ to scale the stochastic term $\sigma_t\delta$, which provides a practical trade-off between sample diversity and geometric sharpness during inference~\cite{dhariwal2021diffusion}.
\subsection{Uncertainty-Aware Robust Consensus Aggregation}
\label{sec:urca}

To improve the stability of multi-sample inference and provide a practical reliability estimate, we unify depth fusion and uncertainty estimation into a robust consensus optimization problem. This module is used as an optional inference-time aggregation scheme that combines stochastic sample consistency with recursive cross-scale consistency from the coarse-to-fine generation process.

\textbf{Multi-sample inference and scale-shift alignment.}
Given an input image $I$, we run the conditional diffusion generator $N$ times to obtain a set of depth predictions $\{\hat{D}^{(n)}\}_{n=1}^{N}$. Since monocular depth predictions are typically comparable only up to an affine transformation (scale and shift), different samples may exhibit global scale-shift ambiguity. To make samples comparable, we align each prediction to a shared reference space:
$
\tilde{D}^{(n)} = \alpha_n \hat{D}^{(n)} + \beta_n ,
$
where $\{(\alpha_n,\beta_n)\}_{n=1}^{N}$ are input-specific affine alignment parameters estimated at inference time (rather than globally learned parameters). We estimate them by minimizing the pairwise alignment energy
\begin{equation}
\label{eq:align_energy}
\mathcal{L}_{\text{align}}(\{\alpha_n,\beta_n\})
=
\sum_{n<m}\left\|\tilde{D}^{(n)}-\tilde{D}^{(m)}\right\|_1
+\lambda\sum_{n=1}^{N}(\alpha_n-1)^2 .
\end{equation}

\textbf{Robust consensus fusion and uncertainty proxy.}
After alignment, we estimate the consensus depth at each pixel by minimizing a unified robust consistency energy. Let $M(x,y)$ denote the consensus depth value at pixel $(x,y)$:
$
M(x,y)=\arg\min_{z}\; E(z;x,y).
$
The energy is defined as
\begin{equation}
\label{eq:consensus_energy}
E(z;x,y)=
\sum_{n=1}^{N}
\rho_s\!\left(
\frac{\tilde{D}^{(n)}(x,y)-z}{\tau_s+\delta}
\right) +\gamma \sum_{k\in\{4,3,2,1\}} w_k\,
\rho_r\!\left(
\frac{\tilde{D}_{g_k}(x,y)-z}{\tau_r+\delta}
\right),
\end{equation}
where $\tilde{D}_{g_k}$ denotes the prediction from recursive generator module $g_k$, first upsampled to the final resolution and then affine-aligned to the same reference space as the stochastic samples; $w_k$ is the scale weight (normalized across levels), $\gamma$ balances the recursive consistency term, $\rho_s$ and $\rho_r$ are robust penalty functions (Charbonnier in our implementation), $\tau_s$ and $\tau_r$ are normalization constants, and $\delta$ is a small numerical stabilization term. The first term enforces stochastic consistency across multi-sample predictions, while the second term enforces recursive consistency along the coarse-to-fine generation trajectory.
We define the pixel-wise {uncertainty proxy} as the minimum residual energy of the consensus optimization:
$
U(x,y)=E(M(x,y);x,y).
$
Thus, the module jointly outputs the robust fused depth map $M$ and the uncertainty proxy map $U$ within the same optimization framework. We emphasize that $U$ is a practical reliability indicator for relative ranking, weighting, or region filtering, rather than a strictly calibrated probabilistic uncertainty.
\textbf{Inference usage.}
In multi-sample inference mode, we use the consensus-fused depth as the final prediction,
$
\hat{D} := M.
$

\vspace{-3mm}
\section{Experiments}
\vspace{-3mm}

\begin{table}[t]
\begin{center}
\resizebox{1\textwidth}{!}{
\begin{tabular}{lllllllllll}
\multicolumn{1}{l}{\bf\ Method}  & \multicolumn{1}{l}
{\cellcolor{pink} Architecture} & \multicolumn{1}{l}
{\cellcolor{pink}AbsRel ↓} & \multicolumn{1}{l}{\cellcolor{pink}Sq Rel↓} & \multicolumn{1}{l}{\cellcolor{pink}RMSE↓} & \multicolumn{1}{l}{\cellcolor{pink}RMSE log↓} & \multicolumn{1}{l}{\bf\cellcolor{blue!30}$\delta < 1.25$\ ↑} & \multicolumn{1}{l}{\bf\cellcolor{blue!30}$\delta < 1.25^2$\ ↑} & \multicolumn{1}{l}{\bf\cellcolor{blue!30}$\delta < 1.25^3$\ ↑}
\\ \hline 
AdaBins \cite{bhat2021adabins} & E-B5+mini-ViT & 0.067 &0.190 &2.960 &0.088 &0.949 &0.992 &0.998
\\
DPT \cite{ranftl2021vision} & VIT-L & 0.060 & - &2.573 &0.092 &0.959 &0.995 &0.996
\\
P3Depth \cite{patil2022p3depth} & ResNet-101 & 0.071 &0.270 &2.842 &0.103 &0.953 &0.993 &0.998
\\
NeWCRFs \cite{yuan2022neural} & Swin-Large & 0.052 &0.155 &2.129 &0.079 &0.974 &0.997 &0.999
\\
BinsFormer \cite{li2024binsformer} & Swin-Large & 0.052 &0.151 &2.098 &0.079 &0.974 &0.997 &0.999
\\
PixelFormer \cite{agarwal2023attention} & Swin-Large & 0.051 &0.149 &2.081 &0.077 &0.976 &0.997 &0.999
\\
VA-DepthNet \cite{liu2023va} & Swin-Large & 0.050 &0.148 &2.093 &0.076 &0.977 &0.997 &0.999
\\
IEBins \cite{shao2023iebins} & Swin-Large & 0.050 &0.142 &2.011 & - &0.978 &0.998 &0.999
\\
iDisc \cite{piccinelli2023idisc} & Swin-Large & 0.050 &0.145 &2.067 &0.077 &0.977 &0.997 &0.999
\\
DCDepth \cite{wang2024dcdepth} & Swin-Large & 0.051 &0.145 &2.044 &0.076 &0.977 &0.997 &0.999
\\
WorDepth \cite{zeng2024wordepth} & Swin-Large & 0.049 & - &2.039 &0.074 &0.979 &0.998 &0.999
\\ \hline
EcoDepth \cite{patni2024ecodepth} &  ViT-L & 0.048 &0.139 &2.039 &0.074 &0.979 &0.998 &1.000
\\  
ZoeDepth \cite{bhat2023zoedepth} &  ViT-L & 0.054 &0.189 &2.440 &0.083 &0.977 &0.996 &0.999
\\
DepthAnything \cite{zhao2024depth} &  ViT-L & 0.046 & - &1.896 &0.069 &0.982 &0.998 &1.000 
\\  \hline

DiffusiongDepth \cite{duan2024diffusiondepth} &Diffusion &0.050 &0.141 &2.016 &0.074& 0.977 &0.998& 0.999
\\  
Repurposing Diffusion \cite{ke2023repurposing} &Diffusion &0.105&- &-& - & 0.904 &- &-
\\ 
DepthFM Diffusion \cite{gui2025depthfm} &Diffusion &0.091&- &-& - & 0.92 &- &-
\\ 
Lotus  \cite{helotus} &Diffusion &0.081&- &-& - & 0.931 &0.987&- \\ 
\hline
DAR-Base \cite{wang2025scalable} &  Autoregressive & 0.046 & 0.114 &1.823 &0.069 &0.985 &0.999 &1.000\\
Ours  &Autoregressive  &0.044 &0.132 &1.712	&0.069&	0.980	&0.997	&0.999
\\ \hline
\end{tabular}}
\end{center}
\vspace{-3mm}
\caption{ A quantitative comparison of the KITTI dataset \cite{geiger2013vision}.}
\label{Kitti_result}
\vspace{-7mm}
\end{table}
\subsection{Implementation Details}
\label{sec:imple}
\textbf{Training setup.}
We use the AdamW optimizer with $(\beta_1,\beta_2)=(0.9,0.95)$ and weight decay $0.05$. The base learning rate is $5\times10^{-5}$ and is linearly scaled by the global batch size divided by $32$.
We apply a 5-epoch linear warm-up followed by cosine decay to $5\times10^{-7}$. 
\textbf{Supervision and diffusion training objective.}
We train the model in a fully supervised manner using ground-truth depth maps, where the supervision is provided by the conditional diffusion objective in continuous latent space (Sec.~\ref{sec:continuous_token}). Specifically, ground-truth depth maps are encoded into scale-wise target depth latents, and each recursive level is supervised by the corresponding conditional diffusion loss.
\textbf{Recursive generation and patch-based context passing.}
The depth representation is generated in a coarse-to-fine recursive manner across four levels. At each level, the previous generator predicts a \textbf{guidance-depth} token, defined as the mean \emph{log-depth} at the current resolution. This token is concatenated to the transformer condition input to provide global scale information before finer structural details are synthesized. For patch-wise generation at finer resolutions, depth maps are divided into square patches. To reduce boundary artifacts, each patch is passed to the next generator stage together with its 4-neighborhood context (top, bottom, left, and right neighboring patches).
\textbf{VCFR and visual backbone settings.}
The proposed VCFR module is compatible with backbones that provide multi-scale visual features. In this work, we evaluate two backbone families: convolutional ResNet backbones~\cite{he2016deep} and Swin Transformers~\cite{liu2021swin}. To strengthen cross-scale feature interaction, we adopt HAHI~\cite{li2023depthformer} as the neck and use an FPN~\cite{lin2017feature} to aggregate multi-scale features into scale-aligned visual conditioning features for recursive refinement. The visual condition dimensionality is set to match the output dimensionality of the neck/FPN stage used for conditioning. The backbone stage channel dimensions are [64, 128, 256, 512] for ResNet and [192, 384, 768, 1536] for Swin.
\textbf{Inference details.}
The uncertainty-aware robust consensus aggregation (URCA, Sec.~\ref{sec:urca}) is not used during training and is applied only in the optional multi-sample inference mode.

\vspace{-1mm}
\subsection{Architecture}
\vspace{-2mm}


\begin{wrapfigure}{r}{0.5\textwidth}
\vspace{-4mm}
\centering
\includegraphics[width=\linewidth]{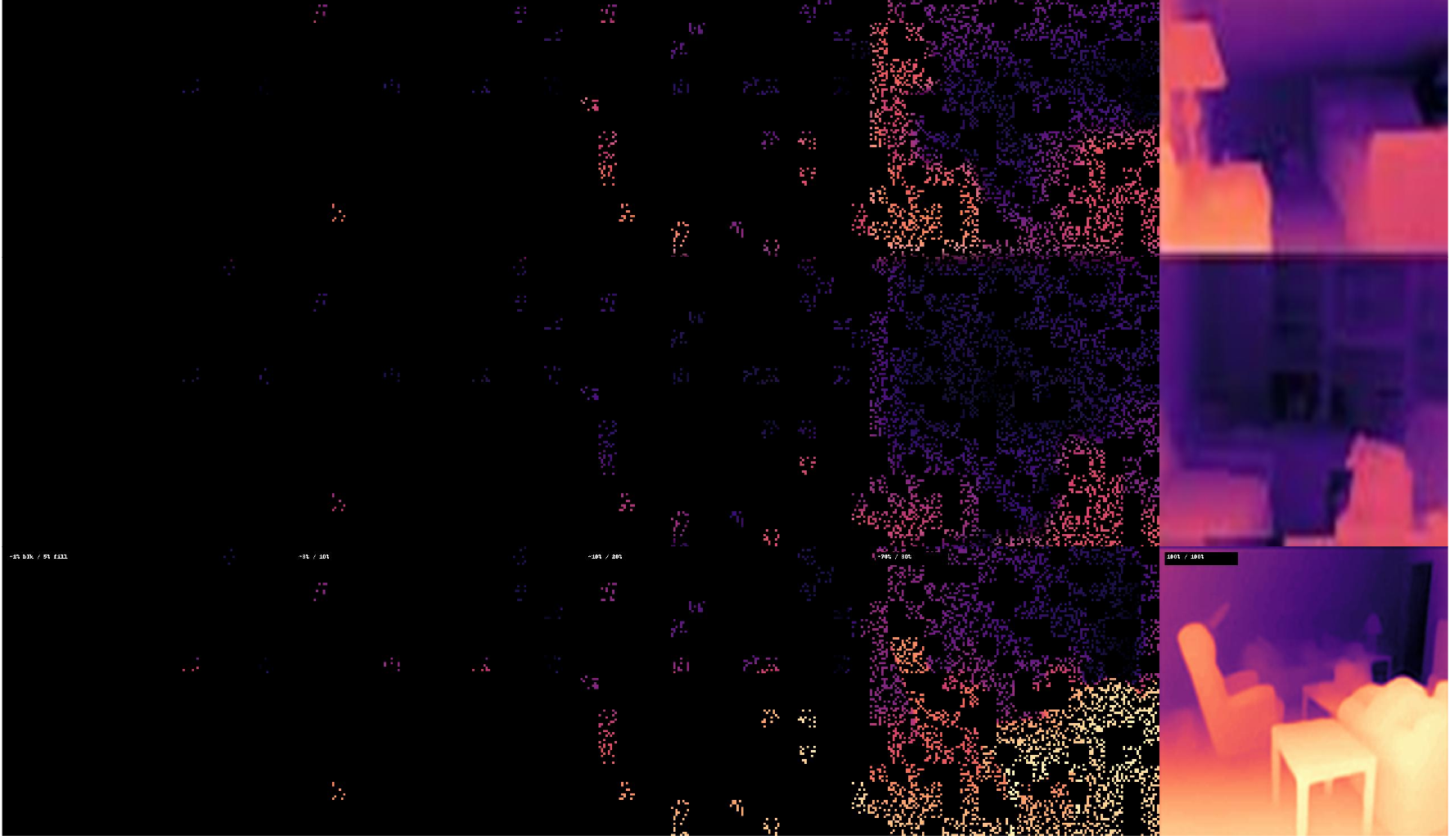}
\vspace{-6mm}
\caption{Different stages of depth in a fractal framework.}
\label{prog}
\vspace{-8mm}
\end{wrapfigure}

We propose a \textbf{Fractal Visual Autoregressive Diffusion Model} that follows a recursive generation strategy. Starting from a coarse latent token, the model progressively generates a high-resolution depth map through multi-stage conditional diffusion and autoregressive prediction. Unlike conventional methods that directly regress depth from RGB inputs, our approach begins with a low-dimensional latent and recursively reconstructs the full depth structure via four hierarchical visual autoregressive diffusion modules. Specifically, the fourth-level module $g_4$ takes a single global latent token (corresponding to a $1\times1$ depth map) as input and predicts the initial coarse depth. Its output serves as a conditional input to the third-level module $g_3$, which takes $16\times$ ($1\times1$) latent tokens and outputs a $4\times4$ resolution depth map, capturing localized structure. The second-level module $g_2$ then takes $4\times4$ patches of $4\times4$ latent tokens (also 16 in total), and predicts a $16\times16$ resolution depth map with more detailed mid-level structure. Finally, the top-level module $g_1$ receives $16\times16=256$ latent tokens, each representing a $16\times16$ region, and recursively performs visual autoregressive diffusion to generate the full-resolution $256\times256$ depth map. This generation process consists of four recursive stages. Each layer outputs a depth representation at a specific resolution, which serves as the conditional input for the next layer. The model thus gradually expands from a global latent to a fine-grained pixel-level depth map. The recursive architecture forms a fractal-like nested hierarchy, where each stage exhibits structural self-similarity and recursively calls. 


\vspace{-1mm}

\begin{table*}[t]
\centering

\begin{minipage}[t]{0.48\textwidth}
\centering
\resizebox{\textwidth}{!}{
\begin{tabular}{lcccc}
\toprule
\textbf{Level} & \textbf{Sequence} & \textbf{Input} & \textbf{Scale} & \textbf{GFLOPs} \\
\midrule
$g_1$ & 256 & $16 \times 16$ & $256 \times 256$ & 26 \\
$g_2$ & 16  & $4 \times 4$   & $16 \times 16$   & 332 \\
$g_3$ & 16  & $1 \times 1$   & $4 \times 4$     & 819 \\
$g_4$ & 1   & $1$            & $1 \times 1$     & 650 \\
\bottomrule
\end{tabular}
}
\caption{Configuration of each level in our fractal visual autoregressive depth estimation framework.}
\label{tab:fractal_config}
\end{minipage}
\hfill
\begin{minipage}[t]{0.48\textwidth}
\centering
\resizebox{\textwidth}{!}{
\begin{tabular}{lccccc}
\toprule
\textbf{Model} & \textbf{Loss} & \textbf{Arch} &
\textbf{Abs\,Rel $\downarrow$} & \textbf{RMSE $\downarrow$} & \textbf{Sq\,Rel $\downarrow$} \\
\midrule
AR  & CrossEnt  & VQ16 & 0.051  & 2.081 & 0.149 \\
AR  & Diff Loss & VQ16 & 0.050  & 2.011 & 0.142 \\ 
VAR & CrossEnt  & VQ16 & 0.051  & 2.044 & 0.145 \\
VAR & Diff Loss & VQ16 & 0.046  & 1.823 & 0.139 \\
VAR & Diff Loss & KL16 & \textbf{0.044} & \textbf{1.712} & \textbf{0.132} \\
\bottomrule
\end{tabular}
}
\caption{Analysis of the diffusion loss on KITTI \cite{geiger2013vision}.}

\label{tab:diffusion_ablation}
\end{minipage}
\vspace{-6mm}
\vspace{-4mm}
\end{table*}
\subsection{Comparisons with Previous Methods}
\begin{figure}[t]
\begin{center}
\includegraphics[width=10cm, height=5cm]{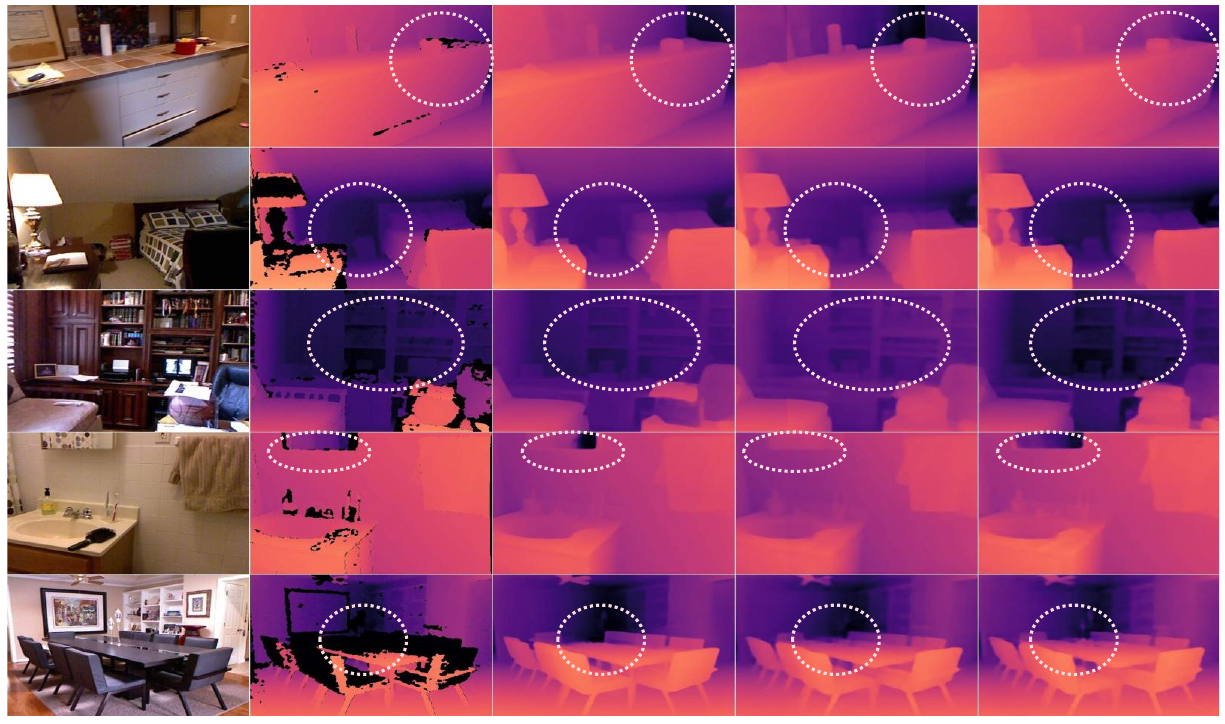}
\end{center}
\vspace{-7mm}
\caption{\textbf{Visual results on NYU.} From left to right:  input images, depth estimation from ground truth, DiffusionDepth \cite{duan2024diffusiondepth}, Repurposing Diffusion \cite{ke2023repurposing}, ours. }
\vspace{-8mm}
\label{nyuvis}
\end{figure}

\vspace{-1mm}
\textbf{Quantitative Results : }
We conduct comprehensive evaluations of the proposed model on the outdoor KITTI dataset \cite{geiger2013vision} and the indoor NYUv2 dataset \cite{silberman2012indoor}. As summarized in Tables \ref{Kitti_result} and \ref{nyu_result}, our method consistently outperforms existing state-of-the-art supervised approaches across multiple evaluation metrics on both datasets. Specifically, we reduce the RMSE to 1.7124 on KITTI and further to 0.197 on NYU.
As illustrated in  Fig. \ref{nyuvis}, our visual autoregressive diffusion framework  captures both local and global structural patterns within the depth distribution. Compared to methods such as Repurposing Diffusion \cite{ke2023repurposing}, DiffusionDepth \cite{duan2024diffusiondepth}, our model demonstrates superior depth estimation quality through a more comprehensive and structured generative process.

\vspace{-5mm}
\subsection{Analysis of the Number of Inference Samples}
\begin{figure}[t]
\centering

\begin{minipage}[t]{0.48\linewidth}
    \centering
    \includegraphics[width=\linewidth,height=3.5cm]{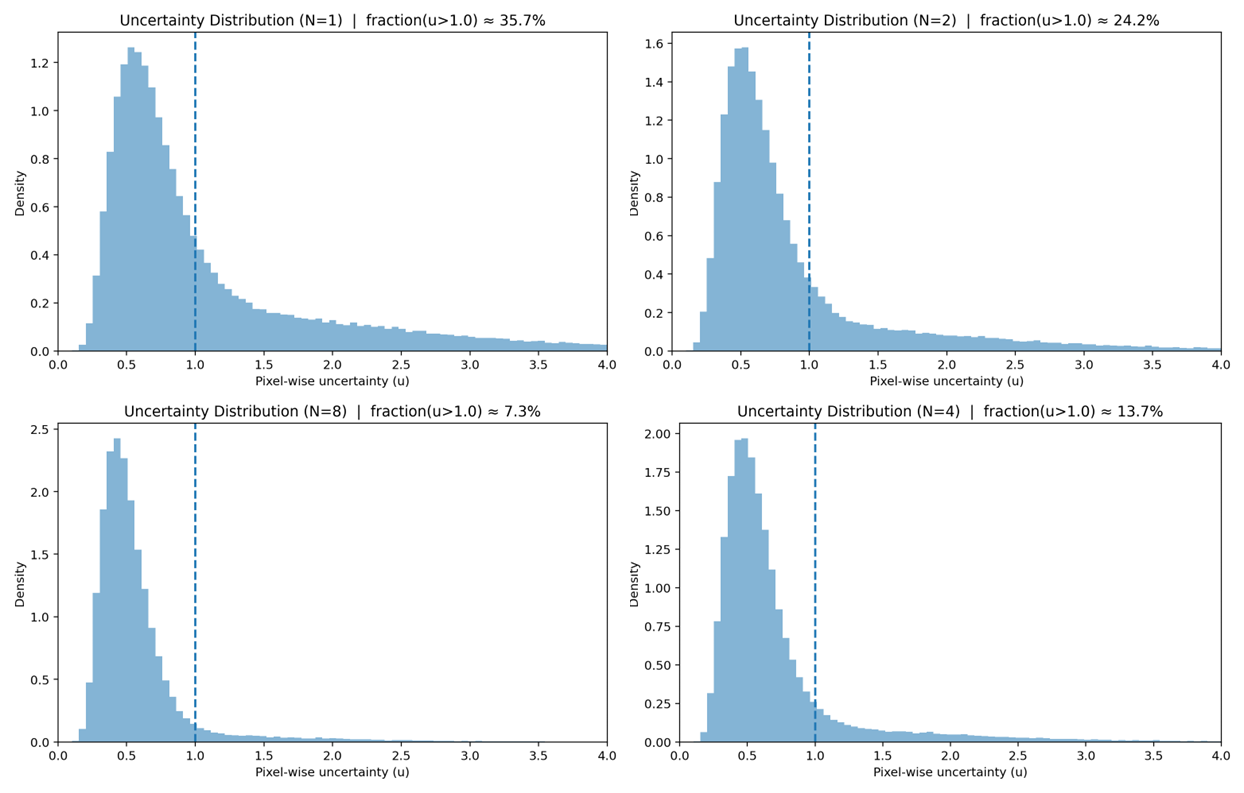}
    \vspace{-6mm}
    \caption{Uncertainty distribution shifts under multi-sample inference with different numbers of runs ($N=1,2,4,8$). In each subplot, the x-axis denotes the pixel-wise uncertainty proxy $u$ and the y-axis denotes density. The dashed vertical line indicates the threshold $u=1.0$.}
    \vspace{-4mm}
    \label{fig:multi_uncert_dist}
\end{minipage}
\hfill
\begin{minipage}[t]{0.48\linewidth}
    \centering
    \includegraphics[width=\linewidth,height=3.5cm]{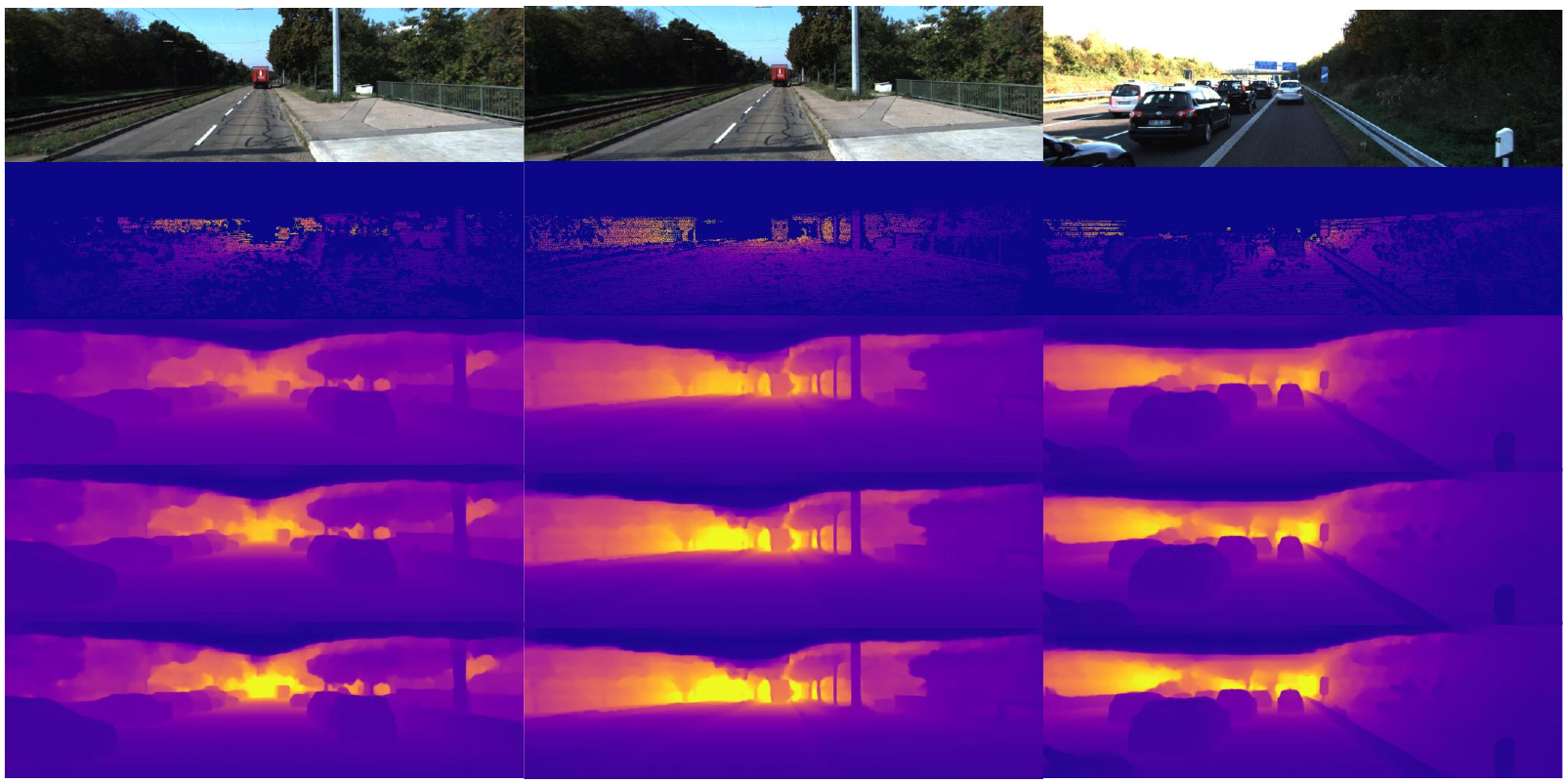}
    \vspace{-6mm}
    \caption{Effect of multi-sample uncertainty aggregation on KITTI. From top to bottom: input RGB images, ground truth, and depth predicted from 1, 4, and 8 diffusion samples.}
  
    \label{fig:multi_qual}
\end{minipage}

\end{figure}

In Table \ref{tab:runs_effect}, We analyze the effect of the number of stochastic inference samples $N$ on model performance  and stability. As $N$ increases, result consistently decrease, while $\delta < 1.25$ consistently increases, indicating that multi-sample inference can steadily improve depth estimation accuracy. Meanwhile, the performance gain exhibits a clear diminishing-return trend: the improvement is more pronounced when increasing from $N=1$ to $N=4$, while further increasing to $N=8$ or $N=10$ still yields gains but with smaller margins. This suggests that a moderate number of samples is sufficient to achieve a favorable performance--cost trade-off.
From the uncertainty distribution perspective in Fig \ref{fig:multi_uncert_dist}, as $N$ increases from 1 to 2, 4, and 8, the distribution of the pixel-wise uncertainty proxy $u$ shifts toward lower values overall, the high-uncertainty tail shrinks noticeably, and the proportion of pixels with $u>1.0$ decreases significantly. This indicates stronger pixel-wise consistency after multi-sample fusion, which is also consistent with the continuous improvement observed in the quantitative metrics.
The qualitative results (Fig.~\ref{fig:multi_qual}) further verify this trend: as the number of samples increases, the predicted depth maps become smoother and more stable in road regions, vehicle contours, and distant structures, with reduced local noise and fewer inconsistent regions, leading to clearer overall geometric structures. Overall, multi-sample inference combined with robust consensus aggregation provides consistent gains in accuracy, stability, and visual quality.

\begin{table}[t]
\begin{center}
\resizebox{0.9\textwidth}{!}{
\begin{tabular}{lllllllllll}
\multicolumn{1}{l}{\bf\ Method}  & \multicolumn{1}{l}
{\cellcolor{pink}Encoder} & \multicolumn{1}{l}
{\cellcolor{pink}AbsRel ↓} & \multicolumn{1}{l}{\cellcolor{pink}Sq Rel↓} & \multicolumn{1}{l}{\cellcolor{pink}RMSE↓} & \multicolumn{1}{l}{\cellcolor{pink}log↓} & \multicolumn{1}{l}{\bf\cellcolor{blue!30}$\delta < 1.25$\ ↑} & \multicolumn{1}{l}{\bf\cellcolor{blue!30}$\delta < 1.25^2$\ ↑} & \multicolumn{1}{l}{\bf\cellcolor{blue!30}$\delta < 1.25^3$\ ↑}
\\ \hline 
AdaBins \cite{bhat2021adabins} & E-B5+mini-ViT & 0.103 &-&0.364 &0.044 &0.903 &0.984 &0.997
\\
P3Depth \cite{patil2022p3depth} & ResNet-101 & 0.104 & -&0.356 &0.043 &0.904 &0.988 &0.998
\\
LocalBins \cite{bhat2022localbins} & E-B5 & 0.099 &- &0.357 &0.042 &0.907 &0.987 &0.998
\\ \hline
NeWCRFs \cite{yuan2022neural} & Swin-Large & 0.095 &0.045 &0.334 &0.041 &0.922 &0.992 &0.998
\\
BinsFormer \cite{li2024binsformer} & Swin-Large & 0.094 &- &0.330 &0.040 &0.925 &0.991 &0.997
\\
PixelFormer \cite{agarwal2023attention} & Swin-Large & 0.090 & -&0.322 &0.039 &0.929 &0.991 &0.998
\\
VA-DepthNet \cite{liu2023va} & Swin-Large & 0.086 &0.043 &0.304 &0.039 &0.929 &0.991 &0.998
\\
IEBins \cite{shao2023iebins} & Swin-Large & 0.087 &0.040 &0.314 &0.038 &0.936 &0.992 &0.998
\\
NDDepth \cite{shao2023nddepth} & Swin-Large & 0.087 &0.041 &0.311 &0.038 &0.936 &0.991 &0.998
\\
DCDepth \cite{wang2024dcdepth} & Swin-Large & 0.085 &0.039 &0.304 &0.037 &0.940 &0.992 &0.998
\\
WorDepth \cite{zeng2024wordepth} & Swin-Large & 0.088 & - &0.317 &0.038 &0.932 &0.992 &0.998
\\ \hline
VPD \cite{zhao2023unleashing} & ViT-L & 0.069 &0.030 &0.254 &0.027 &0.964 &0.995 &0.999
\\
EcoDepth \cite{patni2024ecodepth} &  ViT-L & 0.059 &0.013 &0.218 &0.026 &0.978 &0.997 &0.999
\\  
ZoeDepth† \cite{bhat2023zoedepth} &  ViT-L & 0.077 & - &0.282 &0.033 &0.951 &0.994 &0.999
\\
DepthAnything† \cite{zhao2024depth} &  ViT-L & 0.063 & 0.020 &0.235 &0.026 &0.975 &0.997 &0.999
\\   
VDA-L \cite{chen2025video} &  ViT-L & 0.046& - &- &-&0.978 &- &-
\\  \hline

DiffusiongDepth \cite{duan2024diffusiondepth} &Diffusion &0.085&- &0.295& 0.036 & 0.939 &0.992 &0.999
\\  
Repurposing Diffusion \cite{ke2023repurposing} &Diffusion &0.055&- &-& - & 0.964 &- &-
\\ 
DepthFM \cite{gui2025depthfm} &Diffusion&0.055&- &-& - & 0.963 &- &-
\\ 
Lotus  \cite{helotus} &Diffusion &5.1&- &-& - & 0.97.2 &0.992 &-
\\ \hline
DAR-Base \cite{wang2025scalable} &  Autoregressive & 0.058 & 0.013 &0.214 &0.026 &0.980 &0.997 &0.999
\\ 
Ours  &Autoregressive & 0.049 & 0.011 &0.197 &0.023 &0.984 &0.998 &0.999
\\ \hline
\end{tabular}}
\end{center}
\vspace{-2mm}
\caption{A quantitative comparison on NYU dataset \cite{silberman2012indoor}.}
\label{nyu_result}
\vspace{-10mm}
\end{table}

\begin{figure}[t]
\centering

\begin{minipage}[t]{0.49\linewidth}
    \centering
    \resizebox{\linewidth}{!}{
    \begin{tabular}{ccccc}
    \toprule
    \textbf{Runs (N)} & \textbf{AbsRel $\downarrow$} & \textbf{SqRel $\downarrow$} & \textbf{RMSE $\downarrow$} & \textbf{$\delta < 1.25$ $\uparrow$} \\
    \midrule
    1  & 0.0440 & 0.1320 & 1.712 & 0.9800 \\
    2  & 0.0425 & 0.1296 & 1.702 & 0.9812 \\
    3  & 0.0416 & 0.1282 & 1.696 & 0.9819 \\
    4  & 0.0410 & 0.1272 & 1.692 & 0.9824 \\
    6  & 0.0401 & 0.1258 & 1.686 & 0.9832 \\
    8  & 0.0395 & 0.1248 & 1.681 & 0.9837 \\
    10 & 0.0390 & 0.1240 & 1.678 & 0.9840 \\
    \bottomrule
    \end{tabular}
    }
    \vspace{-3mm}
    \captionof{table}{Effect of the number of inference runs $N$ on depth estimation performance on KITTI \cite{geiger2013vision}.}
    \label{tab:runs_effect}
\end{minipage}
\hfill
\begin{minipage}[t]{0.49\linewidth}
    \centering
    \resizebox{\linewidth}{!}{
    \begin{tabular}{lcccc}
    \toprule
    \textbf{Condition} & \textbf{DSR} & \textbf{Abs Rel.} $\downarrow$ & \textbf{RMSE} $\downarrow$ & \textbf{Cost} $\downarrow$ \\
    \midrule
    Res34+FPN          & $\times2$ & 0.0554 & 2.0502 & 0.60 \\
    Res50+FPN          & $\times2$ & 0.0532 & 1.9724 & 0.74 \\
    MobileNetV3+FPN    & $\times2$ & 0.0492 & 1.9245 & 0.35 \\
    Swin+FPN           & $\times2$ & 0.0458 & 1.8169 & 0.89 \\
    ConvNeXt+FPN       & $\times2$ & 0.0468 & 1.8432 & 1.10 \\
    ConvNeXt+HAHI      & $\times2$ & 0.0448 & 1.7680 & 1.24 \\
    Swin+HAHI          & $\times2$ & \textbf{0.0440} & \textbf{1.712} & 1.00 \\
    Swin+HAHI          & $\times4$ & 0.0443 & 1.7280 & 0.98 \\
    \bottomrule
    \end{tabular}
    }
    \vspace{-3mm}
    \captionof{table}{Comparison of visual conditions on KITTI \cite{geiger2013vision}. Cost denotes normalized inference cost,where the highest-cost setting is normalized to 1. }
    \label{tab:visual}
\end{minipage}

\vspace{-4mm}
\end{figure}




\begin{figure}[t]
\begin{center}
\includegraphics[width=10cm, height=4cm]{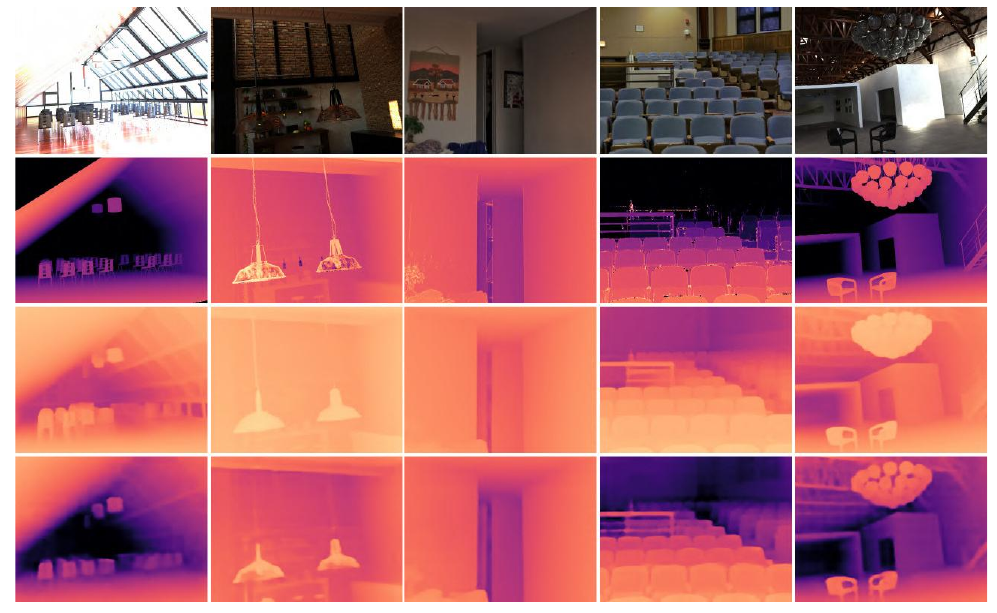}
\end{center}
\vspace{-7mm}
\caption{\textbf{Visual results.} From top to bottom; input images, ground truth, depth from discrete VQ tokens, and depth from continuous KL tokens.. 
}
\vspace{-5mm}
\label{lianvsli}
\end{figure}

\vspace{-4mm}
\subsection{Module Design and Analysis}
\vspace{-2mm}

\textbf{Diffusion Loss:}The results in Table~\ref{tab:diffusion_ablation} show the effectiveness of the diffusion for monocular depth estimation under different autoregressive settings.
First, under an AR backbone with a VQ tokenizer, we treat the continuous latents before the VQ layer as diffusion tokens and replace the original cross-entropy objective with the diffusion objective, which improves depth estimation metrics. Under the VAR backbone, the RMSE is further reduced. These consistent gains across the evaluated AR variants suggest that the diffusion objective provides a more effective supervision signal for depth prediction than discrete token likelihood training in our setting.
Second, after introducing the diffusion objective, replacing the VQ-16 tokenizer with a KL-16 tokenizer further reduces the RMSE to 1.712. This result indicates that diffusion-based continuous modeling benefits from the richer continuous structure preserved by KL codes. As illustrated in Fig.~\ref{lianvsli}, continuous depth tokens produce reconstructions that are visually closer to the ground truth than discrete VQ-16 tokens. The discrete formulation shows noticeable quantization artifacts (e.g., staircase effects on large planar regions) and less stable transitions around thin structures, whereas continuous KL-16 tokens yield smoother depth gradients while better preserving sharp boundaries and fine geometric details. These qualitative results further support the advantage of continuous latent-space modeling for fine-grained depth geometry.


\textbf{Analysis of Visual Conditions.}
As shown in Table \ref{tab:visual}, although visual conditions such as MobileNetV3+FPN and ResNet34+FPN achieve significantly faster inference speeds compared to Swin+HAHI, they suffer from noticeable drawbacks in accuracy. Conversely, high-accuracy models like ConvNeXt+FPN attain performance that is comparable to or slightly better than Swin, but incur much higher computational cost and inference latency. In contrast, Swin+HAHI strikes a favorable balance between accuracy and efficiency, ranking among the top in multiple evaluation metrics. As a result, we adopt it as the default visual condition configuration in our model.
In addition, we evaluate encoder-decoder structures based on depth latent spaces with down-sampling rates of ×4 and ×2. While the lower-resolution latent space (×4) offers slight advantages in runtime, the higher-resolution (×2) consistently yields better accuracy in depth prediction. Therefore, we select the ×2 setting, considering the trade-off between computational efficiency and predictive performance.

\vspace{-4mm}
\subsection{Inference Latency}
\vspace{-4mm}
\begin{table}[h]
\centering
\resizebox{1\textwidth}{!}{
\begin{tabular}{lcccccccc}
\toprule
{Method} & {Marigold} & {Lotus} & {E2E-Mono} &
{Monodepth2} & {MonoViT} & DiffusiongDepth & {Jasmine}& Ours \\
\midrule
MACs    & 133T & 2.65T & 2.65T & 21.43G & 25.63G & - & 35.7G& 132G \\
Runtime & 9.88s & 157ms & 152ms & 33ms & 29ms & 223ms & 172ms& 64ms \\
\bottomrule
\end{tabular}
}
\caption{MACs and runtime on input resolution $1024\times320$.}
\vspace{-8mm}
\label{tab:macs_runtime}
\end{table}

As shown in Table~\ref{tab:macs_runtime}, our method offers a favorable accuracy--efficiency trade-off: while it is slower than lightweight CNN-/Transformer-based modles (e.g., Monodepth2 and MonoViT), it is substantially faster than diffusion-based approaches (e.g., Marigold, DiffusionDepth, and Jasmine) under the same input resolution, while maintaining stronger depth estimation performance.
\vspace{-2mm}
\subsection{Ablation Study}
\begin{wraptable}{r}{0.45\textwidth}
\vspace{-2mm}
\centering
\resizebox{0.45\textwidth}{!}{
\begin{tabular}{llllllllll}
{\cellcolor{pink}AR} & 
\multicolumn{1}{c}{\cellcolor{pink}VAR} & 
\multicolumn{1}{c}{\cellcolor{pink}Diffusion} & 
\multicolumn{1}{c}{\cellcolor{pink}Fractal} & 
\multicolumn{1}{c}{\cellcolor{pink}VCFR} & 
\multicolumn{1}{c}{\cellcolor{blue!30}Cost} & 
\multicolumn{1}{c}{\cellcolor{blue!30}AbsRel $\downarrow$} & 
\multicolumn{1}{c}{\cellcolor{blue!30}Sq Rel $\downarrow$} & 
\multicolumn{1}{c}{\cellcolor{blue!30}RMSE $\downarrow$} & 
\multicolumn{1}{c}{\cellcolor{blue!30}$\delta < 1.25$ $\uparrow$} 
\\ \hline 
\checkmark  &  & & &  & 1.000 & 0.083 & 0.039 & 0.314 & 0.938
\\ \hline 
& \checkmark &  &  &  & 0.017 & 0.079 & 0.037 & 0.279 & 0.949
\\ \hline 
& \checkmark & \checkmark &  &  & 0.720 & 0.063 & 0.020 & 0.235 & 0.975
\\ \hline  
& \checkmark & \checkmark & \checkmark &  & 0.045 & 0.058 & 0.013 & 0.212 & 0.982
\\ \hline 
& \checkmark & \checkmark & \checkmark & \checkmark & 0.051 & 0.049 & 0.011 & 0.197 & 0.984
\\ \hline  
\end{tabular}
}
\vspace{-2mm}
\caption{Ablation study of the framework on the NYU dataset.}
\vspace{-7mm}
\label{tab:ablation_result_model}
\end{wraptable}

We conduct ablation studies on the NYUv2 dataset to evaluate the contribution of each component, with results summarized in Table~\ref{tab:ablation_result_model}. Here, cost denotes the normalized inference cost, with the AR baseline normalized to 1.000 (lower is better). Therefore, each value represents the relative inference overhead compared with the AR model under the same evaluation setting.
\textbf{Visual Autoregressive Structure.} Comparing the AR and VAR variants (first two rows), the AR baseline shows limited depth estimation accuracy and a high inference cost. Replacing AR with a VAR formulation improves performance while substantially reducing the normalized cost from 1.000 (AR baseline) to 0.017, indicating that scale-wise autoregressive generation is more suitable for dense depth prediction.
\textbf{Conditional Denoising Diffusion Objective.} We then introduce a conditional denoising diffusion objective to model depth latents in continuous space and alleviate limitations introduced by discrete tokenization. As shown in Table~\ref{tab:ablation_result_model}, adding diffusion significantly improves depth estimation accuracy. However, the multi-step denoising process increases the normalized inference cost to 0.720, motivating a more efficient generator organization.
\textbf{Fractal Structure.}
 To reduce the cost introduced by diffusion-based scale-wise generation, we adopt the proposed fractal recursive design, which composes the model from reusable small-scale autoregressive modules. This design markedly improves the accuracy-cost trade-off, reducing the normalized cost to 0.045 while further improving prediction performance.
\textbf{Visual-Conditioned Feature Refinement}
Finally, we add the VCFR module to enhance visual-depth conditioning at each recursive scale. As shown in Table~\ref{tab:ablation_result_model}, VCFR yields the best overall performance across all metrics while maintaining a low normalized cost of 0.051, validating its effectiveness for cross-modal alignment and fine-grained geometric refinement.

\vspace{-2mm}
\section{Conclusion}
We proposed a fractal visual autoregressive diffusion framework for monocular depth estimation. Our method reformulates depth prediction as a coarse-to-fine next-scale autoregressive process and organizes the generators in a recursive fractal hierarchy, improving the computational structure of dense prediction. The proposed VCFR module enhances RGB--depth alignment at each scale, while a conditional diffusion objective models depth latents in continuous space without explicit discretization. We further introduce an uncertainty-aware robust consensus aggregation scheme for multi-sample inference to improve prediction stability and provide a practical reliability proxy. Experiments on standard benchmarks show strong performance and validate the effectiveness of the proposed design.

\vspace{-1mm}

%
%
\bibliographystyle{splncs04}
\bibliography{main}
\end{document}